\documentclass{article}
\usepackage{spconf,amsmath,graphicx}
\usepackage{multirow}%表格
\usepackage{booktabs}
\usepackage{threeparttable}
\usepackage{fancyhdr}
\usepackage{enumitem}
\setlist{nosep, leftmargin=14pt}

\usepackage{mwe} % to get dummy images
\usepackage{amsfonts}
\title{SurgPLAN: Surgical Phase Localization Network for Phase Recognition}
\name{Xingjian Luo$^{1}$\textsuperscript{*}, You Pang$^{1,2}$\textsuperscript{*}, Zhen Chen$^{1}$\textsuperscript{$\dagger$}, Jinlin Wu$^{1}$\textsuperscript{$\dagger$}, Zongmin Zhang$^{1}$, Zhen Lei$^{1}$, Hongbin Liu$^{1}$ \thanks{* Equal Contribution} \thanks{$\dagger$ Correspoonding Authors}
}
\address{$^{1}$Centre for Artificial Intelligence and Robotics (CAIR), HKISI-CAS, Hong Kong SAR, China \\
$^{2}$The Hong Kong Polytechnic University, Hong Kong SAR, China}
\begin{document}
%\ninept
%
\maketitle

% their predictions are unstable, indicating that the predicted outcome within one phase will be turbulent. 
% predictions based on each frame lead to unstable results within one stage. is turbulent and will cause performance degradation.
%
\begin{abstract}
Surgical phase recognition is crucial to providing surgery understanding in smart operating rooms. Despite great progress in automatic surgical phase recognition, most existing methods are still restricted by two problems. First, these methods cannot capture discriminative visual features for each frame and motion information with simple 2D networks. Second, the frame-by-frame recognition paradigm degrades the performance due to unstable predictions within each phase, termed as \textit{phase shaking}. To address these two challenges, we propose a \underline{Surg}ical \underline{P}hase \underline{L}oc\underline{A}lization \underline{N}etwork, named SurgPLAN, to facilitate a more accurate and stable surgical phase recognition with the principle of temporal detection. 
Specifically, we first devise a Pyramid SlowFast (PSF) architecture to serve as the visual backbone to capture multi-scale spatial and temporal features by two branches with different frame sampling rates. 
Moreover, we propose a Temporal Phase Localization (TPL) module to generate the phase prediction based on temporal region proposals, which ensures accurate and consistent predictions within each surgical phase.
Extensive experiments confirm the significant advantages of our SurgPLAN over frame-by-frame approaches in terms of both accuracy and stability.
 
\end{abstract}
\section{Introduction}
\label{sec:intro}
Surgical phase recognition plays an important role in the comprehensive understanding of surgery, especially in monitoring surgical procedures \cite{panesar2020promises} and coordinating surgical teams \cite{kennedy2020computer}. In particular, the cataract surgery performed on eyeballs demands careful operations \cite{dataset} and will benefit from accurate surgery phase recognition to a great extent.

% Cataracts can cause varying degrees of vision loss and have become one of the leading risks for blindness. A common way to address the risk of blindness associated with cataracts is to conduct the cataract surgery. As the surgery is performed on the eyeballs, it is necessary to operate carefully and accurately, and the success of the operation is highly dependent on the surgeon's skill and proficiency. Therefore, training junior surgeons to perform cataract surgery also takes a lot of effort.

% Over the past few years, computer-assisted techniques have made significant contributions to the development of intelligent medicine. A key technique in computer-assisted surgery is surgical workflow recognition. Surgical phase recognition plays an important role in operation assessment and surgical training of cataract surgeries. By automatically identifying the stage of the surgery, it can generate operational descriptions of the corresponding stage, which can be further used for surgical procedure review, operational skill assessment, and so on. According to the assessment results, surgeons can optimize their surgical practices and avoid potential risks in future surgeries, which will greatly improve the success and safety of cataract surgery. In addition, this can assist in the surgical training of junior surgeons to reduce the workload of other surgeons.

\begin{figure}[t]
% \begin{minipage}[b]{1\linewidth}
  \centering
  \centerline{\includegraphics[width=0.49\textwidth]{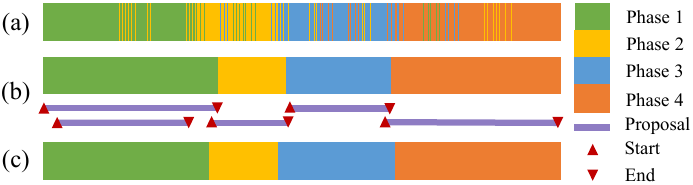}}
%  \vspace{1.5cm}
% \end{minipage}
\vspace{-0.5em}
\caption{In surgical phase recognition, (a) the frame-by-frame prediction-based methods have the \textit{phase shaking} problem. In contrast, (b) the detection-based paradigm generates continuous phase predictions with temporal region proposals, which is more consistent with (c) the ground truth phases.}
\label{fig1}
\vspace{-1em}
\end{figure}

Recently, many deep learning methods have been proposed for automatic surgical phase recognition \cite{deepphase}\cite{tecno}\cite{svrcnet}\cite{trans-svnet}. After extracting the features of each surgical video frame with 2D convolutional neural networks (CNNs), these methods customized different architectures to exploit the temporal knowledge of surgical videos, \textit{e.g.}, the temporal convolution \cite{tecno}, long short-term memory (LSTM) \cite{svrcnet} and transformer \cite{trans-svnet}. Nevertheless, these methods struggle to produce stable and continuous surgical stage predictions, limited by two problems. On the one hand, simple 2D CNNs cannot capture enough temporal features for adjacent frames, especially at different temporal scales. While certain methods also utilize 3D CNNs \cite{tran2015learning} to capture both spatial and temporal features, they are inefficient and computationally consuming. On the other hand, these methods formulated frame-by-frame surgical phase predictions are not holistic and lack the global field of information, thereby prone to the \textit{phase shaking} problem, \textit{i.e.}, the predicted phase labels are not consistent inside one surgical phase, as illustrated in Fig.~\ref{fig1} (a). This problem can significantly impact the performance of phase prediction and, more critically, impede the efficiency of surgeons.
%. The incorrect predictions are interspersed within the same phase as illustrated in Fig.~\ref{fig1} (a). This problem can significantly impact the performance of phase prediction and, more critically, impede the efficiency of surgeons.

%framework
\vspace{-0.5em}
\begin{figure*}[t!]
\centering
\includegraphics[width=0.9\textwidth,keepaspectratio=false]{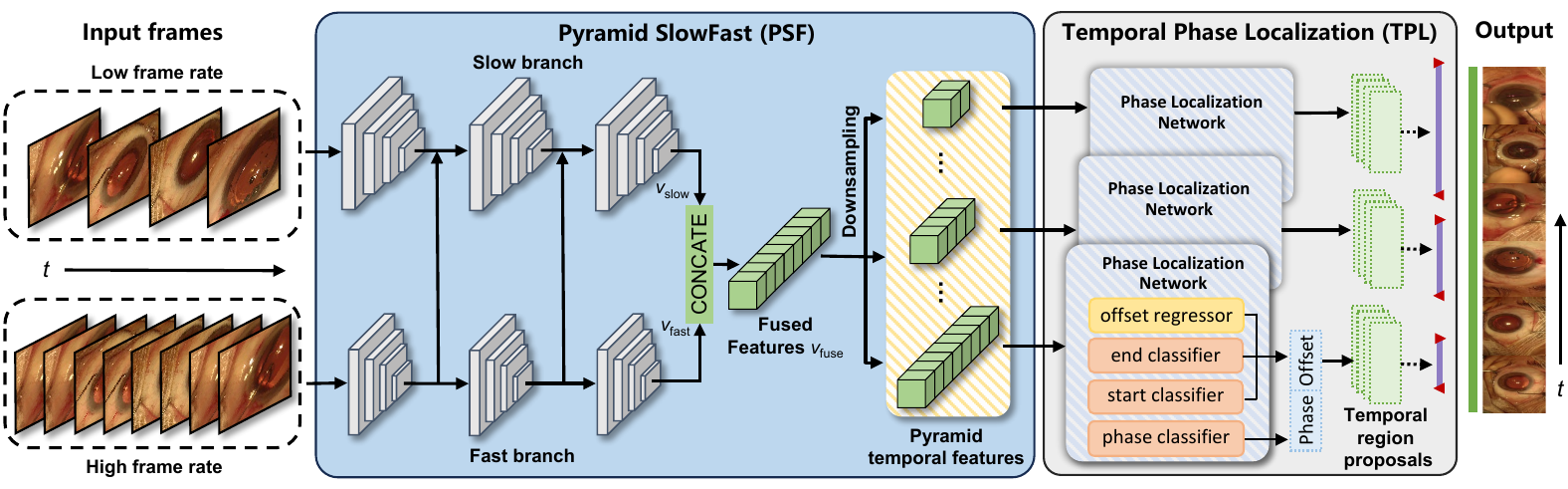}
\vspace{-0.5em}
\caption{The SurgPLAN framework. In particular, the PSF module first captures discriminative spatial and temporal information across two branches with different frame sampling rates, and then the TPL module generates surgical phase period predictions from temporal region proposals based on the detection paradigm.}
\label{fig:framework}
\end{figure*}

To address these two problems, we propose a novel SurgPLAN framework to achieve accurate and stable surgical phase recognition, including a Pyramid SlowFast (PSF) architecture to extract multi-scale spatial and temporal features and a Temporal Phase Localization (TPL) module to perform surgical phase prediction with temporal region proposals. Specifically, instead of simply employing a 2D CNN to extract the visual features of each frame, the PSF adopts a slow branch and a fast branch to process the input surgical video with different frame sampling rates, thereby capturing multi-scale features containing a more comprehensive temporal knowledge for accurate phase recognition. Moreover, to address the \textit{phase shaking} problem in frame-by-frame recognition, the TPL module first generates temporal region proposals with the estimations of phase start-end location, and its class given the center point and then selects the optimal proposals as the phase prediction for surgical videos, as elaborated in Fig. \ref{fig1} (b). It is worth noting that our SurgPLAN framework is partially inspired by the temporal action localization approaches in natural scenarios \cite{actionformer}\cite{tridet}\cite{lin2019bmn}. The basic idea behind these methods is detecting the boundary of the actions in an untrimmed video and generating proposals for action retrieval. To validate the effectiveness of our SurgPLAN framework, we conduct extensive experiments on the public cataract surgery dataset with classification and detection metrics, confirming the advantage over state-of-the-art methods in terms of both accuracy and stability.

\section{METHODOLOGY}
\label{sec:format}
\subsection{Overview}% 目标是xx，给定sequence，网络，输出。

Previous frame-by-frame prediction methods \cite{deepphase}\cite{tecno}\cite{svrcnet} are prone to predict intermittent wrong phases during one phase due to the independence of each prediction. Our SurgPLAN aims to generate temporal region proposals, including a start-end point with a phase class and its score. 
Specifically, by clearly demarcating the start-end point, a phase segment is delineated that is immune to the discontinuous problem, thereby enhancing the stability and accuracy of the prediction.
The SurgPLAN first takes a video as input and extracts features with spatial and temporal information through the PSF. Then, the TPL module utilizes these features to generate temporal region proposals. Finally, the SurgPLAN selects appropriate proposals as phase predictions of surgical videos.

\vspace{-1em}
\subsection{Pyramid SlowFast Architecture}
\label{ssec:subshead}
% 介绍字符
% slow原理
% fast原理
% 融合
% 加个公式 f_i = .....
% 介绍公式
% pretraining 好处，原理
\noindent \textbf{Spatial Temporal Feature Extraction.}
The PSF utilizes the SlowFast \cite{slowfast} on untrimmed videos $V\in \mathcal{R}^{T \times (H\times W) \times N}$ to extract spatial and temporal information where $T$, $H$, $W$, $N$ refers to the video frame length, height, width, and RGB channel. First, we get the Slow Path feature $ v_{\rm slow} \in \mathcal{R}^{T \times \gamma C }$ and Fast Path feature $ v_{\rm fast} \in \mathcal{R}^{T \times C }$, as follows:
\vspace{-0.5em}
\begin{equation}
\begin{split}
    v_{\rm slow} &= \phi_{\rm slow} (V), \\  
    v_{\rm fast} &= \phi_{\rm fast} (V), \\
    v_{\rm fuse} &= \mathcal{D}([v_{\rm slow}, v_{\rm fast}]),
\end{split}
\end{equation}
where $\phi_{\rm slow}$ and $\phi_{\rm fast}$ refer to the corresponding branch networks that generate features originally intended to be fed into the classification head, and $T$, $C$, and $ \gamma$ refer to the video frame length, feature channel size, and feature channel ratio. After that, we transform the concatenated features $ [v_{\rm slow}, v_{\rm fast}] \in \mathcal{R}^{T \times (1+\gamma) C }$ to the feature space through a downsampling backbone $\mathcal{D}$. 
%The final feature $ v_{\rm fuse}\in \mathcal{R}^{T^{\star} \times C^{\star}} $ contains both spatial and temporal information with an adequate temporal length $T^{\star}$ and channel dimension $C^{\star}$. 

\noindent \textbf{Pyramid Temporal Features.}
These extracted frame-by-frame features $v_{\rm fuse}$ are compressed multiple times by max-pooling to reduce their temporal length, as follows:
\begin{equation}
\begin{split}
    \text{{Pyramid}}(v_{\rm fuse}) = \{\text{Max}(v_{\rm fuse},S_{\rm window})\},\\
    {\rm w.r.t. \,\,} S_{\rm window} \in \{1, 2,4\},
\end{split}
\end{equation}
where $S_{\rm window}$ is the Pooling window with kernel and stride sizes of 1, 2, and 4. The pyramid method generates 3 sequences of features with different downsampling rates.
These features capture discriminative spatial and temporal information at different scales, thus facilitating a better adaptation to the specific temporal segment size required in the TPL. 

\vspace{-1em}
\subsection{Temporal Phase Localization Module}\label{ssec:subhead}
We modify the existing structure \cite{actionformer}\cite{tridet}, by adapting and refining it to be more appropriate for the surgical environment that tends to have prolonged periods of action and less noticeable variances in the foreground among features.

\vspace{1pt}
\noindent \textbf{Temporal Proposal.}
The features $f \in \text{{Pyramid}}(v_{\rm fuse}) $ represent one sequence of temporal-spatial features at a certain scale. $f^i$ refers to features at the temporal position $i$ passed through three classification neural networks and one regression neural network.
\begin{equation}
\begin{split}
    p_{\rm phase}^{i} &= \psi_{\rm phase}(f^{i}), \\
    p_{\rm start}^{i} &= \psi_{\rm start}(f^{i}), \\
    p_{\rm end}^{i} &= \psi_{\rm end}(f^{i}), 
\end{split}
\end{equation} 
$\psi_{\rm phase}$, $\psi_{\rm start}$ and $\psi_{\rm end}$ are classification neural networks $\psi$ with the same structure but different classification heads at the last output layer corresponding to their output dimensions. They independently predict surgical phase probability 
$p_{\rm phase}^{i}$, and start-end point probabilities $p_{\rm start}^{i}$, $p_{\rm end}^{i}$ of feature point $f^i$. The structure of $\psi$ is
\begin{equation}
f_{l}^{i} = \sigma(\mathcal{N}(\mathrm{FC}(f_{l-1}^{i}))),
\end{equation}
where $\sigma$ is the activation function, $\mathcal{N}$ is the layer normalization function, $\mathrm{FC}$ refers to the fully-connected layer, $i$ is the temporal position of the processed feature $f$, and $l$ represents which layer it is in the hidden layers. 

The regression network shares a similar backbone structure as the classifier $\psi$ but has a different output head. It outputs a conditional probability distribution 
$P(\mathcal{B}|f^i)=
\{\ {p^t} \mid t\in \{i-\frac{B}{2},i+\frac{B}{2}\} \ \text{and} \ t \neq i\ \}$
where each $p^t$ in the temporal interval represents the probability of being the start-end point of the segment given the target feature point $f^i$ being the center point which is located at $i$, and $B$ denotes the bin size.
For each region proposal in TPL, given the center point $f^i$, the start point $\hat{t}_{\rm start}(f^i)$ and end point $\hat{t}_{\rm end}(f^i)$ are predicted by the corresponding start and end subset of probability distribution $P(\mathcal{B}|f^i)$ and classification feature points $p_{\rm start}^{[i-\frac{B}{2}: i-1]}$ or $p_{\rm end}^{[i+1:i+\frac{B}{2}]}$ from $\psi_{\rm start}$ and $\psi_{\rm end}$ as follows: 
\begin{equation} \label{eq:5}
\begin{split}
    \hat{t}_{\rm start}( f^i )&= \mathrm{Argmax}(P(\mathcal{B}|f^i)^{[i-\frac{B}{2}:i-1]}+p_{\rm start}^{[i-\frac{B}{2}:i-1]}), \\
    \hat{t}_{\rm end}(f^i )&= \mathrm{Argmax}(P(\mathcal{B}|f^i)^{[i+1:i+\frac{B}{2}]}+p_{\rm end}^{[i+1:i+\frac{B}{2}]}),
\end{split}
\end{equation}
where $\mathrm{Argmax}$ is the function that identifies the temporal position in a sequence that produces the maximum value. Therefore, given the center feature, we add up two sequences of probabilities and select the position with the highest probability being the start and end points. 

Regarding selecting center feature points that meet the requirements for training, we first determine whether the feature point is within a ground truth segment. 
Secondly, we evaluate its distance to the segment edges. If it is too close to an edge, it may not exhibit characteristics of the center point. If it is far from both edges, meaning the segment length is long, the bin can not fully cover the whole segment at the current scale. By considering both scale rate and bin size, we choose those adequate points as training center point samples.

\vspace{1pt}
\noindent \textbf{Proposal Selection.}
We use temporal non-maximization suppression (NMS) to reduce the number of highly overlapped or low-score region proposals. For region proposals at different scales, we multiply the reverse of the downsampling rate to retrieve the original length. The remaining non-overlapping proposals are the final predictions of each phase in the video.

\begin{figure}[t]
% \begin{minipage}[b]{1\linewidth}
  \centering
  \centerline{\includegraphics[width=0.5\textwidth]{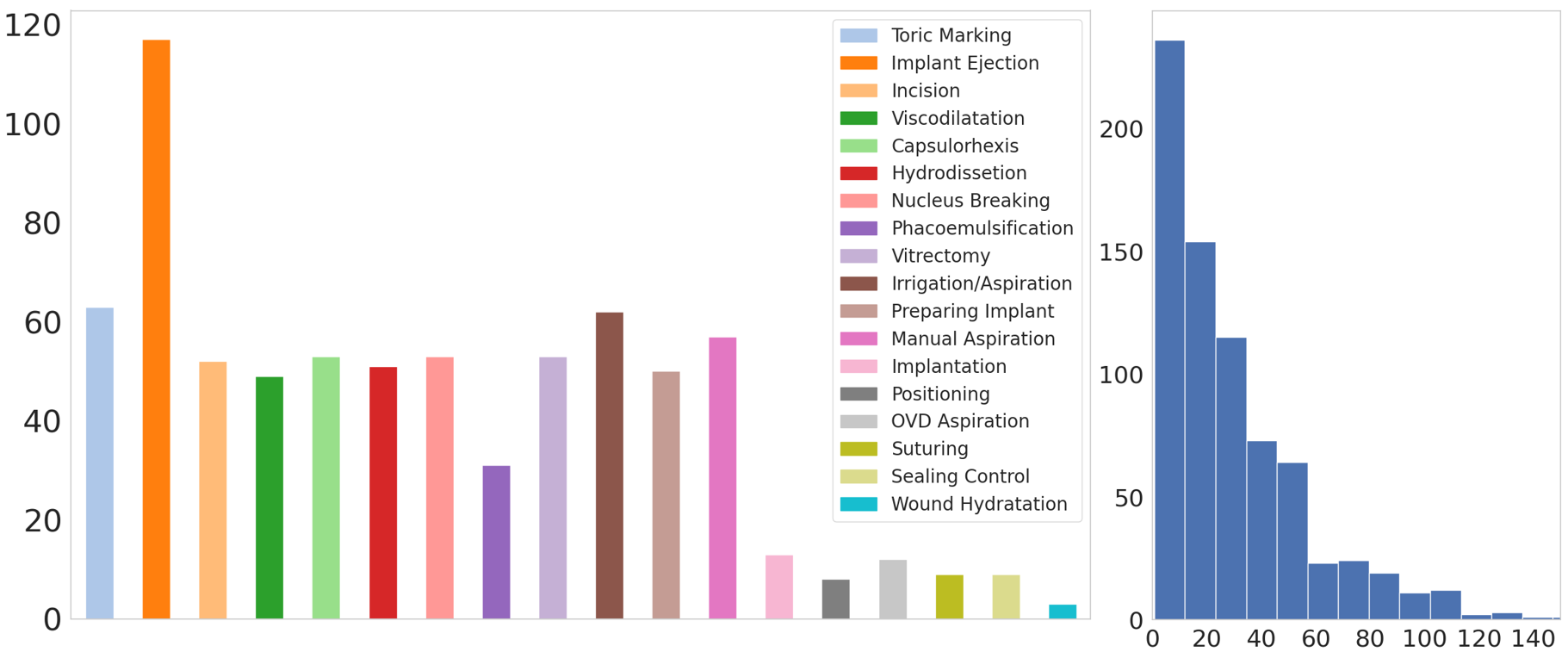}}
%  \vspace{1.5cm}
 % \centerline{(a) Segment duration distribution}\medskip
% \end{minipage}
% \begin{minipage}[b]{1\linewidth}
%   \centering
%   \centerline{\includegraphics[width=6.0cm,height=4.0cm]{label_count.png}}
% %  \vspace{1.5cm}
  % \centerline{(a)                         (b)}\medskip
% \end{minipage}
\caption{The statistics of the CATARACTS dataset, including (a) segment counts of phases and (b) duration distribution of phase segment (in seconds) .}
\label{fig:seg}

\end{figure}

\vspace{1pt}
\noindent \textbf{Loss Functions.}
For the classification task, the original videos are filled with a vast amount of background features, so we utilize the Focal Loss \cite{focal} as $L_{\rm focal} = (1-p_{\rm phase}^{i})^{\gamma}\log(p_{\rm phase}^{i})$, where $p_{\rm phase}^{i}$, $\gamma$ refers to the phase probability and focusing parameter. It allows the model to focus more on the foreground rather than the background. 
For the segment regression loss, calculated jointly from the start-end features, and offsets, we adapted the Intersection over Union (IoU) Loss into the one-dimensional temporal scenario. 
\begin{equation} \label{eq:4}
\begin{split}
   L_{\rm IoU} = 1- \underbrace{\frac{\mathrm{min}(\hat{s}_{\rm start},s_{\rm start})+\mathrm{min}(\hat{s}_{\rm end},s_{\rm end})}{\mathrm{max}(\hat{s}_{\rm start},s_{\rm start})+\mathrm{max}(\hat{s}_{\rm end},s_{\rm end})}}_{\rm IoU},
\end{split}
\end{equation}
where $s_{\rm start}$, $s_{\rm end}$, and $\hat{s}_{\rm start}$, $\hat{s}_{\rm end}$ refer to the ground truth and predicted distance between the center point and the start or end point. The temporal IoU is calculated by dividing the union of the ground truth segment and the predicted segment by the intersection of two segments on the temporal dimension. The temporal IoU loss liberates the model from being confined to frame-by-frame predictions, instead allowing it to anticipate surgical stages from a global perspective. Therefore, our SurgPLAN can predict each surgical stage with greater accuracy and precision.

%作用： ground truth

\section{EXPERIMENTS}\label{sec:pagestyle}
\subsection{Experimental Settings}
\noindent \textbf{CATARACTS Dataset.} We conduct our experiment on the public CATARACTS dataset \cite{dataset} and follow the standard split to divide 25 cataract surgery videos for training and the remaining 25 videos for the test. This dataset contains 19 phase categories, including one background category without a clear surgical purpose. We transform the frame-by-frame labels into segments of surgical phases, and each segment consists of a start time, an end time, and a phase label. After the process, the dataset contains 1,536 non-intersected phase segments, and 745 of them are non-background phases, which mainly range from 0 to 60 seconds. The segment duration and label count distribution are shown in Fig. \ref{fig:seg}.

\noindent \textbf{Implementation Details.} We conduct the surgical visual pre-training of the PSF on the training set. 
%We first resize the frames into the shape of $224\times224\times3$. Then we take 32 frames as an input sequence with a step size of 1 and 4 at the slow and fast path and set the initial learning rate as 0.1 with a cosine scheduler and a Stochastic Gradient Descent (SGD) optimizer during the training. The training process is conducted for 130 epochs. After feature extraction, we set the stride rate as 30 to shorten these feature sequences to an adequate length. 
The frames are first resized to the shape of $224\times224\times3$. Then, we select 32 frames as the input sequence with step sizes of 1 and 4 for the slow and fast paths, respectively. The feature dimensions of the Slow and Fast branches are 2048 and 256. For the visual pre-training of the PSF, the initial learning rate is set as 0.1 with a cosine scheduler, and a Stochastic Gradient Descent (SGD) optimizer is used.

For the training of the TPL, the learning rate is 0.001, with 10 warm-up epochs and 100 training epochs. We use the Adam optimizer with a momentum of 0.9. The downsampling rates in the PSF are 1, 1/2, and 1/4. The corresponding ranges for selecting center points are (0, 8), (4, 32), and (32, 200). The bin size is set as 24. 

%For instance, at the original scale, the target feature point must be more than 0 and less than 8 units away from the beginning and end edges; in other words, the point should be inside one segment; meanwhile, the total length should not exceed 16 units. After downsampling by half, we aim to filter those points that are more than 4 units and less than 32 units away from each side of the edge, which means the corresponding segment length must be at least 16 and less than 128 units when projecting to the original size. 

%Through the implementation of these filters, we want to enable the model to learn segments of appropriate length that correspond to each scale. And the final range is (0,8),(4,32), and (32,200) with 1, 1/2, and 1/4 of its original scale.

In the inference stage, to enhance the generalization ability and also avoid losing potentially correct temporal region proposals, we use the Soft-NMS  \cite{softnms} for the TPL and set a confidence score threshold as 0.2. We operate two approaches to classify the phase of each segment by calculating the average feature phase score and center point feature score.

\begin{table}[t]
    \begin{center}
            \caption{Comparison with classification metrics.}
        \resizebox{0.45\textwidth}{!}{
        \begin{tabular}{c| c c c c} \hline 
        \toprule[1pt]
            Method& Accuracy& F1& Recall&Precision \\
            \hline
            ResNet \cite{resnet}&$59.50$&$42.48$&$46.48$&$42.42$\\
            TeCNO \cite{tecno}& $62.94$&$48.13$&$50.06$&$49.24$\\
            SlowFast \cite{slowfast}&$67.51$&$63.71$&$67.96$&$65.57$\\
            SV-RCNet \cite{svrcnet}&$72.11$&$58.73$&$60.65$&$62.33$\\
            SurgPLAN & $\boldsymbol{83.10}$&$\boldsymbol{71.87}$&$\boldsymbol{72.83}$&$\boldsymbol{73.13}$\\\hline
            \toprule[1pt]
        \end{tabular}
        }
        \label{tab1}
    \end{center}
    \vspace{-2em}
\end{table}

\noindent \textbf{Evaluation Metrics.} We utilize both detection and classification based metrics to evaluate surgical phase recognition. As for detection, we employ the mean Average Precision (mAP)\footnotemark to indicate the Intersection over Union (IoU) compared with ground truth segments. For classification, we follow the standard evaluation protocols and adopt the accuracy, F1 score, precision, and recall to perform the comprehensive evaluation. The higher values of these metrics indicate superior performance in surgical phase recognition.
\footnotetext{The mAP is calculated by taking the average of nine equally spaced nodes ranging from 0.1 to 0.9.}

%\vspace{-1em}
% 对比SOTA，层次
\subsection{Comparisons with State-of-the-art Methods}
\label{ssec:subhead}
To validate the effectiveness, we compare our SurgPLAN with state-of-the-art methods of surgical phase recognition. As shown in Table \ref{tab1}, TeCNO \cite{tecno} achieves an accuracy of 62.94\%, 
SlowFast \cite{slowfast} reaches an accuracy and F1 score of 67.51\% and 63.71\%, 
and SV-RCNet \cite{svrcnet} has an accuracy of 72.11\% and F1 of 58.73\%. In contrast, our SurgPLAN achieves the best performance with 83.10\% in accuracy and 71.87\% in F1, outperforming the SV-RCNet \cite{svrcnet} with the advantage of 13.14\% in F1. 
Compared with SlowFast \cite{slowfast}, our SurgPLAN can improve both accurateness and stability by 15.59\% in accuracy and 8.16\% in F1.
Note that DeepPhase \cite{deepphase} validated at the simpler version of the CATARACTS dataset with only 14 phases annotated, obtains an inferior accuracy of 78.28\%. Therefore, these results confirm the superiority of our SurgPLAN in surgical video analysis.

% We have evaluated our method from various perspectives to compare it with frame-wise methods. For frame-wise methods, due to the presence of some erroneous classifications within a phase segment that prevents the formation of a complete segment with clear start and end points, we utilized the frame-wise accuracy to compare other frame-wise methods(DeepPhase \cite{deepphase}, TeCNO \cite{tecno}) on the CATARACTS dataset \cite{dataset} only.
% In the experiments that are conducted on the same CATARACTS testing dataset, 
% We fine-tuned a ResNet \cite{resnet} as the baseline of frame-wise phase prediction whose accuracy is 66.58\% and F1 score is 49.69\%. The TeCNO \cite{tecno} network reaches an accuracy of 62.94\% and an F1-score of 48.13\%. 

% And the DeepPhase \cite{deepphase}, which was performed at the same CATARACTS video but in a previous dataset version that only contains 14 phase labels that annotated by themselves, has an accuracy of 78.28\% of Long Short Term Memory(LSTM) backbone and an accuracy of 74.92\% of Gated Recurrent Unit(GRU) backbone. All these methods inevitably faced the problem of generating non-continuous phase predictions. 

% Compared with previous methods, our method achieved the highest accuracy and F1-score of 83.10\% and 71.87\% compared with all the frame-wise models on the same CATARACT Dataset while generating uninterrupted smooth phase predictions, see Table \ref{tab1} and figure \ref{fig:fig3}. The experiment shows that using a temporal localization model can eliminate the discontinuity problem and enhance stability and accuracy.

\begin{figure}[t]
\begin{minipage}[b]{1\linewidth}
  \centering
  \centerline{\includegraphics[width=0.99\textwidth]{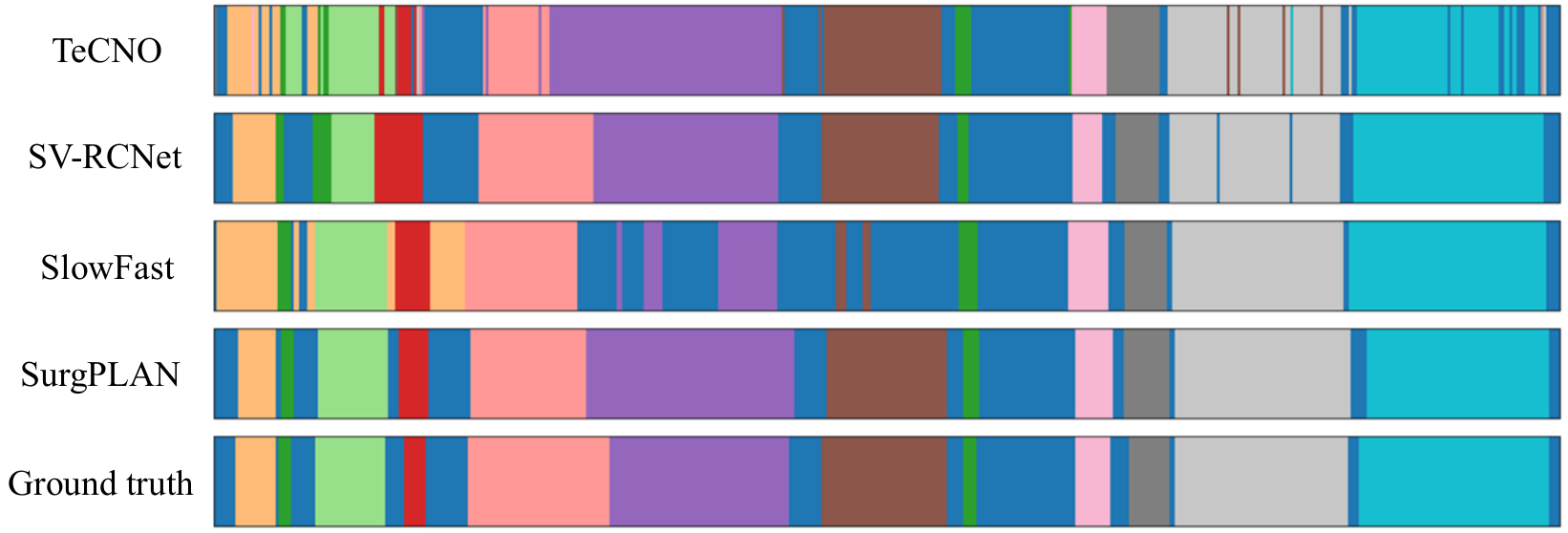}}
%  \vspace{1.5cm}
\end{minipage}
% \centering
% \includegraphics[width=0.5\textwidth]{compare_pred.pdf}
\caption{Color-coded ribbon comparison. Each color represents one phase label.}
\label{fig:fig3}
\end{figure}

We further qualitatively compare the color-coded ribbon results. As shown in Fig.~\ref{fig:fig3}, our SurgPLAN outperforms TeCNO \cite{tecno}, 
SlowFast \cite{slowfast}, 
and SV-RCNet \cite{svrcnet}, and is the closest to ground truth. Note that our SurgPLAN solves the \textit{phase shaking} problem remarkably. In this way, these results confirm the temporal localization in SurgPLAN can eliminate the discontinuity with more accurate predictions.

% superiority of our SurgPLAN in surgical video analysis.

%\vspace{-1em}
%对比实验目的 temporal 对比

\subsection{Ablation Study}
\label{ssec:subhead}
To investigate the impact of visual backbone design, we perform a comparison with different designs, as shown on Table \ref{tab2}. 
Compared with untrained backbones, the trained one has a higher mAP of 59.36\% and 69.79\% in both aggregations. This result reveals that a proficient backbone yields superior outcomes in subsequent TAL tasks.
The center aggregation outperforms the average aggregation by 10.43\% and 8.86\% in both trained and untrained scenarios. This suggests that the features within a segment could potentially be diluted by insignificant features in the boundary values, and the mid-point values ensure that the most salient features are captured, thereby enhancing prediction accuracy. 

\begin{table}[t]
    \begin{center}
    \caption{Comparison of visual backbones for SurgPLAN.}
        \resizebox{0.45\textwidth}{!}{
        \begin{tabular}{c c | c c c c}
            \hline
            \toprule[1pt]
            train & Agg. & AP$_{10}$ & AP$_{50}$ &AP$_{90}$& mAP\\ %\footnotemark\\
            \hline
             &avg&$36.05$&$29.03$&$3.76$&$24.59$\\ 
             &center&$45.66$&$36.13$&$10.04$&$33.45$\\ 
            \checkmark&avg&$79.69$&$65.73$&$15.57$&$59.36$ \\
            \checkmark&center&$\boldsymbol{87.61}$&$\boldsymbol{74.78}$&$\boldsymbol{32.54}$& $\boldsymbol{69.79}$\\ 
            \toprule[1pt]
        \end{tabular}
        }
        %{\footnotesize \item[*] The mAP is calculated by taking the average of nine equally spaced nodes ranging from 0.1 to 0.9.}
        \label{tab2}
    \end{center}
    \vspace{-1.5em}
\end{table}
% \footnotetext{The mAP is calculated by taking the average of nine equally spaced nodes ranging from 0.1 to 0.9.}
\vspace{-0.5em}
\section{CONCLUSION}
\label{sec:typestyle}
In this work, we propose a SurgPLAN based on the detection paradigm to facilitate surgical phase recognition. The SurgPLAN first captures discriminative spatial and temporal features through dual branches with different frame sampling rates and then predicts surgical phase segments from temporal region proposals. Extensive experiments confirm that our SurgPLAN remarkably outperforms existing frame-based methods by alleviating the \textit{phase shaking} problem.

\vfill
%\pagebreak


\bibliographystyle{IEEEbib}
\bibliography{refs}

\end{document}